\def\year{2023}\relax
\documentclass[letterpaper]{article} 
\usepackage{aaai23}  
\usepackage{times}  
\usepackage{helvet}  
\usepackage{courier}  
\usepackage[hyphens]{url}  
\usepackage{graphicx} 
\usepackage{natbib}  
\usepackage{caption} 
\DeclareCaptionStyle{ruled}{labelfont=normalfont,labelsep=colon,strut=off} 
\usepackage{algorithm}
\usepackage{algorithmic}
\usepackage{multicol}
\usepackage{multirow}
\usepackage[thinlines]{easytable}
\usepackage{booktabs}
\usepackage{array}
\usepackage{makecell}
\usepackage{subfigure}
\usepackage[dvipsnames]{xcolor}
\newcolumntype{L}[1]{>{\raggedright\let\newline\\\arraybackslash\hspace{0pt}}m{#1}}
\newcolumntype{C}[1]{>{\centering\let\newline\\\arraybackslash\hspace{0pt}}m{#1}}
\newcolumntype{R}[1]{>{\raggedleft\let\newline\\\arraybackslash\hspace{0pt}}m{#1}}

\usepackage{newfloat}

\usepackage{listings}
\floatstyle{ruled}
\newfloat{listing}{tb}{lst}{}
\floatname{listing}{Listing}
\nocopyright
\usepackage{subfiles}
\usepackage{xcolor}
\usepackage{makecell}
\usepackage{soul}
\usepackage{amsmath}
\title{Do LLMs Find Human Answers To Fact-Driven Questions Perplexing? \\  A Case Study on Reddit}
\author{
    Parker Seegmiller, Joseph Gatto, Omar Sharif\\ 
    Madhusudan Basak, Sarah Masud Preum
}
\affiliations{
    Dartmouth College Department of Computer Science\\

    pkseeg.gr@dartmouth.edu
}

\begin{document}

\maketitle

\begin{abstract}


Large language models (LLMs) have been shown to be proficient in correctly answering questions in the context of online discourse. However, the study of using LLMs to model \textit{human-like} answers to fact-driven social media questions is still under-explored. In this work, we investigate how LLMs model the wide variety of human answers to fact-driven questions posed on several topic-specific Reddit communities, or subreddits. We collect and release a dataset of 409 fact-driven questions and 7,534 diverse, human-rated answers from 15 r/Ask\{Topic\} communities across 3 categories: profession, social identity, and geographic location. We find that LLMs are considerably better at modeling highly-rated human answers to such questions, as opposed to poorly-rated human answers. We present several directions for future research based on our initial findings.
\end{abstract}

\section{Introduction}
Large language models (LLMs) have been used for several social computing tasks, such as sentiment analysis \cite{deng2023llms}, content moderation \cite{kolla2024llm}, and question answering \cite{xiong-etal-2019-tweetqa} on social media, with varying degrees of success. It is important to characterize the extent to which LLMs are in line with human preferences on such tasks. This characterization involves several subtasks, including identifying social media scenarios in which LLMs can generate human-like content, assessing the factuality of LLM-generated content in such scenarios, and determining whether LLMs'  capacity to model online discourse is in line with human preferences. 

Through the lens of exchanging fact-driven information on social media, we examine a vast  source of social media question-answering data, namely 15 r/Ask\{Topic\} communities on Reddit, such as r/AskMen and r/AskNYC. Reddit is a pseudo-anonymous social media website, enabling users to share rich personal content on a wide variety of topics. Users can participate in several communities, each devoted to one specific topic, including complex and uncertain topics. Specifically, r/Ask\{Topic\} communities allow millions of users to pose and answer topic-specific questions to a community of peers interested in those topics. 

\begin{figure}[t]
\centering
\includegraphics[width=0.8\linewidth]{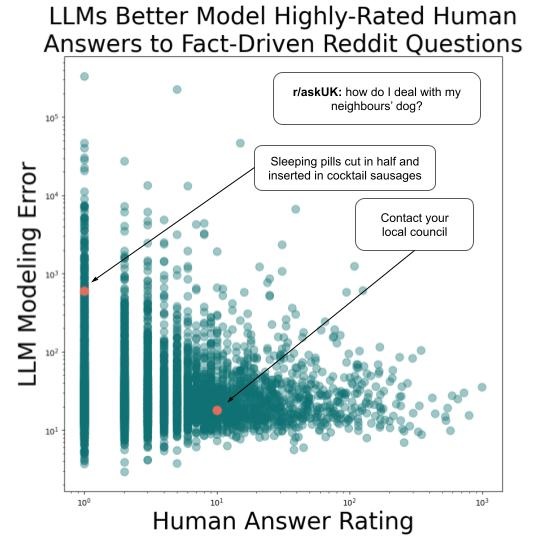} 
\caption{LLM (fine-tuned Sheared LLaMA 1.3B) modeling error (perplexity) and human ratings of 7,534 human answers to fact-driven questions posed on Reddit's r/Ask\{Topic\} communities in log scale. In general, LLM modeling and human perception are well-aligned. As an example, see the two divergent answers to the question asked on r/UK. The LLM assigns low perplexity to the highly-rated human answer, and higher perplexity to the low-rated human answer.}.
\label{fig:preview}
\vspace{-20pt}
\end{figure}

In this study, we are interested in whether LLMs can model \textit{fact-driven} questions posed on these subreddits. As a motivating example, consider this question from r/AskHistorians: ``Why did Spain not join either the Axis or Allied powers in WWII?'' Despite having a factually-grounded, historical answer, there is a wide range of plausible human answers to this question, all of which can be considered ``correct''. Prior work has explored the capacity of LLMs to answer fact-driven questions correctly through autoregressive generation \cite{wang2023survey}. Under-explored, however, is analysis of LLM capacity to intrinsically model the breadth of human answers to fact-driven questions. In this work, we thus aim to answer questions such as (i) Do LLMs find human-written answers to fact-driven questions perplexing? and (ii) Does LLM perception of human-written answers correlate with human ratings?  Answering such questions will scope the capacity of LLMs as social media text generation agents from a novel perspective.  We investigate these questions through the lens of Reddit's r/Ask\{Topic\} subreddits and make the following contributions.

\begin{table}[t]
\scriptsize
\centering
\begin{tabular}{L{1.3cm}|l|L{3.9cm}}
    \textbf{Subreddit} & \textbf{Objectivity} &  \textbf{Question} \\\hline
     Historians & Fact-Driven & Why did Spain not join either the Axis or Allied powers in WWII? \\
     Academia & Subjective & Should I do my PhD and Postdoc in the same university?   \\\hline
     OldPeople & Fact-Driven & How different were Bars/Pubs before smart phones?\\
     WomenOver30 & Subjective & Who makes between \$70-80K? Age, occupation and how you got there.  \\\hline
     UK & Fact-Driven & Why do we have smaller houses than most of the world? \\
     NYC & Subjective & Does the awesome city make you care less about the weather?   \\\hline
\end{tabular}
\caption{Examples of Fact-Driven and Subjective Questions from the r/Ask\{Topic\} Subreddit Communities.}
\label{tab:examples}
\end{table}

\begin{enumerate}
    \item We develop a novel generalizable framework to characterize how LLMs model the wide variety of human responses to fact-driven questions on Reddit. 
    \item We curate a dataset of 409 fact-driven questions and 7,534 answers, along with 494 and 8,177 non-fact-driven questions and answers, gathered from 15 of Reddit's largest r/Ask\{Topic\} communities covering a wide variety of profession, geographic location, and social identity topics.
    \item We provide an intrinsic analysis of different LLM settings on this task by comparing LLM perplexity and human answer ratings, finding that LLMs are considerably better at modeling highly-rated human answers, as opposed to poorly-rated human answers. We also discuss implications of these findings for future research.
\end{enumerate}

Our analysis framework, dataset, and results provide a rich source of hypothesis-generating resources for downstream NLP, social science, and web-based research. 

\section{Related Works}

\subsection{Social Media Question Answering}
Despite being a well-explored research question, there is limited focus on question answering (QA) using social media data, attributed to its distinctive challenges, including hashtags, typos, platform-specific phenomena, and the informal nature of the text \cite{10.1145/3560260}. \citet{xiong-etal-2019-tweetqa} develop a question-answering dataset by crawling tweets and writing question-answer pairs from the tweet. Other researchers have explored Reddit question and comment answer threads for relevant tasks \cite{doi:10.1177/0196859919865250, hew2024examining}.  Recent studies have demonstrated the effectiveness of LLMs in solving the QA task \cite{kamalloo-etal-2023-evaluating}. \citet{staab2023beyond} showed that LLMs can infer many private details about Reddit users via posts and comments in r/Ask subreddits, indicating the need for broader privacy protections beyond LLM text memorization. \citet{kolla2024llm} illustrated how LLMs are bad at specific rules-related content moderation using data from five different subreddits. However, there is a limited exploration into how effectively LLMs model human responses to questions posed on social media platforms.


\subsection{Factual Question Answering}
The key attribute of a QA dataset is its communicative intent: information-seeking and probing. Information-seeking questions aim to obtain factual responses while probing questions are more subjective and context-dependent. While most QA research emphasizes generating accurate factual responses across diverse domains \cite{zhang-etal-2023-survey-efficient}, limited attention has been paid to modeling the responses generated by QA models. \citet{lin-etal-2022-truthfulqa} introduced TruthfulQA to assess the likelihood of models replicating human falsehoods. Their findings revealed that language models generate false responses resembling misconceptions to deceive humans. The advent of LLMs significantly transformed the QA paradigm due to their capacity to offer creative responses to questions. Therefore, it is crucial to understand how much LLM-generated responses model humans' answers. To address this, we present a new dataset and investigate how LLMs model a wide variety of human answers to fact-driven questions posted on Reddit.


\section{Background and Methodology}


\subsection{Definitions}
\noindent
\textbf{Fact-Driven Questions}
We are particularly interested in determining how LLMs model human answers to \textit{fact-driven} questions on social media. By fact-driven questions, we mean posts which contain a question with factually-grounded answers. We consider each post on Reddit's r/Ask\{Topic\} communities to either be fact-driven, or \textit{subjective} (non-fact-driven), as exemplified in Table \ref{tab:examples}. While this binary formulation of fact-driven is a simplification of a complex idea, classifying posts in this manner enables us to filter out questions which would likely require human experience to answer. 

\noindent
\textbf{Answer Score}
Each comment is assigned, via Reddit, a score based on peer engagement and human preference. This peer-assigned score is a measure of the upvote to downvote ratio normalized by comment age.\footnote{https://www.reddit.com/wiki/faq/} We use this measure to proxy peer perception. Highly-rated answers are deemed more valuable or helpful, while low-rated ones are seen as less helpful, aligning with prior research on social media engagement \cite{sharma2020engagement, trunfio2021conceptualising}. 


\noindent
\textbf{Perplexity}
To determine whether an LLM models a human answer well, we use perplexity \cite{jelinek1977perplexity}. Intuitively, perplexity measures how ``surprised'' the LLM is by an answer. If the perplexity is low, the LLM is expecting to see this answer, meaning the LLM can model the autoregressive linguistic properties of the answer. This metric is commonly used to measure performance of language models in encoding linguistic phenomena \cite{belinkov2019analysis}. We use perplexity to measure how well an LLM models human answers to fact-driven questions asked on Reddit's r/Ask\{Topic\} subreddits. Given a human answer $a$ comprised of a sequence of English tokens $a = [a_1, \cdots, a_{|a|}]$, an LLM $M$ assigns autoregressive conditional token probabilities $p_M(a_i | a_1 \cdots a_{i-1})$ to each token $a_i$ in answer $a$. The perplexity of an answer $a$ under language model $M$ is defined as follows. 

   
\begin{gather*}
    PPL_M(a) = exp\{ - \frac{1}{|a|} \sum_i^{|a|} \log p_M(a_i | a_1 \cdots a_{i-1} ) \}
\end{gather*}

Perplexity is the exponentiated average negative log-likelihood of tokens in a sequence, and can be thought of as a measure of LLM modeling error over a sequence.

\subsection{Relevant Data Collection and Processing}

Some of the largest communities on Reddit are the r/Ask\{Topic\} communities, including millions of members who engage in topic-specific discussions with peers, e.g., r/AskWomen has 5.5 million members. We collect fact-driven question-answers from these communities using the following steps.

\textbf{First}, 15 r/Ask\{Topic\} communities across three categories are selected for investigation, as seen in Table \ref{tab:dataset}. We select these popular subreddits to capture diverse content, enabling us to investigate fact-driven questions with high peer engagement (captured as numbers of upvotes and comments).

\textbf{Second}, a random sample of 50\% of all posts and comments, dating back to 2011, are collected from each of these communities. This represents several million such posts and comments. We filter to exclude deleted or removed posts and comments, or exceptionally long posts and comments ($> 10,000$ characters) since they often contain personal narratives rather than fact-driven questions. 
    
\textbf{Third}, posts with less than three top-level comments are then excluded, as we wish to examine questions with high peer engagement. We consider top-level comments to be answers to the primary question in each post.

\begin{table}[t]
\small
\centering
\begin{tabular}{ll|ccc}
    \textbf{Category} & \textbf{Type} &\textbf{\#Q} & \textbf{ \#A} & \textbf{\%F} \\\hline
   \multirow{5}{*}{Profession} & Historians & 35 & 164 & 42\% \\
     & Academia & 13 & 72 & 37\% \\
    & Engineers & 27 & 394 & 42\% \\
    & Culinary & 42 & 302 & 45\% \\
    & Photography & 21 & 168 & 44\% \\\hline
   \multirow{5}{*}{Identity} & Men & 15 & 332 & 50\% \\
    & Women & 100 & 2925 & 47\% \\
   & MenOver30 & 7 & 180 & 32\% \\
   & WomenOver30 & 9 & 268 & 31\% \\
  & OldPeople & 7 & 245 & 58\% \\\hline
    \multirow{5}{*}{Geographic} & UK & 67 & 1684 & 48\% \\
     & NYC & 49 & 392 & 51\% \\
    & France & 7 & 273 & 47\% \\
     & Singapore & 8 & 132 & 47\% \\
    & Argentina & 2 & 11 & 50\% \\\hline
\end{tabular}
\caption{Final dataset statistics of the questions and answers gathered from 15 r/Ask\{Topic\} subreddits across 3 topic categories. Here, \#Q, \#A, and \%F indicate the number of fact-driven questions, the number of answers to those questions, and the percentage of filtered questions which were deem fact-driven.}
\label{tab:dataset}
\end{table}

\begin{figure*}[htbp]
\centering
\includegraphics[width=1.8\columnwidth]{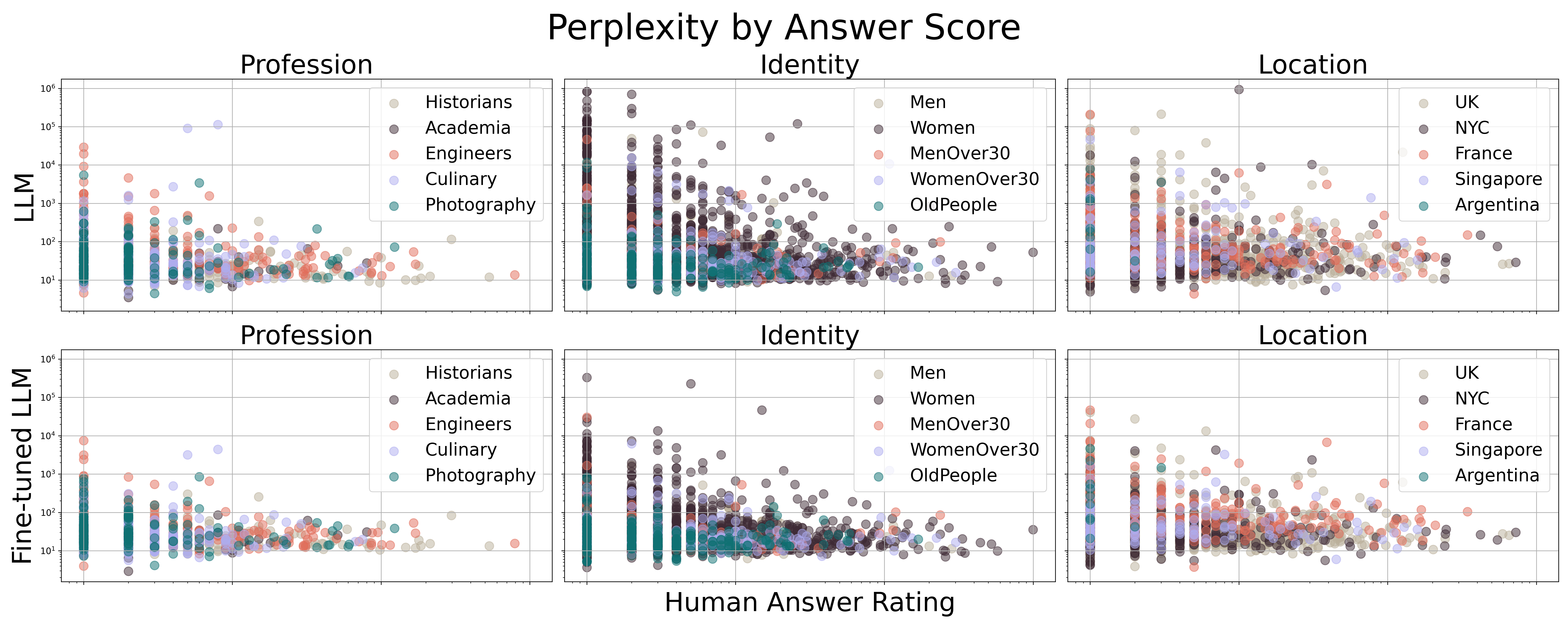} 
\caption{LLM perplexity of answers to fact-driven questions posed on 15 of Reddit's r/Ask\{Topic\} communities, compared with the peer-assigned score of answers. The perplexities of the top row are calculated by the vanilla \textit{SL} LLM, and the bottom row are calculated the fine-tuned \textit{SLFT} LLM. For each graph, the X and Y axes indicate peer-assigned scores and LLM's perplexity in log scale, respectively.}
\label{fig:main}
\vspace{-10pt}
\end{figure*}

\textbf{Fourth}, we then exclude posts whose comments all have the same rating. As we wish to explore LLM modeling capability as it correlates with human perception, we are primarily interested in questions whose answers have varying human ratings. When top-level comments all have the same rating, this typically indicates that the post received less peer engagement and is therefore less suitable for our study. The filtering thus far results in a sample of 903 questions (posts) and 15,711 answers (top-level comments). 

\textbf{Finally}, we filter to consider only fact-driven questions. We do this by passing each question to the large, open-source LLaMA-2-70B model \cite{touvron2023llama}, asking whether the post contains a question with a factually-grounded answer. Filtering social media posts using LLMs is common in social media analysis \cite{kikkisetti2024using, gatto2024scope}. To verify that LLaMA-2-70B performs this fact-driven question binary classification task intuitively, we select a sample of 100 posts and perform inter-annotator agreement, with each post being labeled by two annotators. Each annotator is familiar with Reddit and has vast experience in social computing and factuality assessment tasks. Two human annotators agree with each other on 83\% of posts. Both human annotators agree with LLaMA-2-70B on 41\% of posts, and at least one agreed on 58\% of posts. We attribute these inter-annotator agreement scores to the limitation of treating questions as either fully fact-driven or fully subjective, when in reality there exists a range of subjectivity. However, we deem this classification to be a better alternative to considering \textit{all} posts, as such posts vary widely in context and scope. 

This process results in a dataset of 409 \textit{questions} (posts containing fact-driven questions) from 15 r/Ask\{Topic\} across 3 categories, and 7,534 \textit{answers} (top-level comments containing answers) to those questions. We release all filtered questions and answers, including those that are not deemed fact-driven by LLaMA-2-70B, and leave further analysis of this filtering for future work.

\subsection{Models}

To evaluate LLM modeling of answers to questions posed on Reddit's r/Ask\{Topic\} subreddits, we utilize Sheared LLaMA, a light-weight version of the open-source LLaMA2-7B LLM which is pruned to only 1.3 billion parameters using target structured pruning \cite{xia2023sheared}. We employ Sheared Llama as opposed to it's larger counterpart Llama2-7B as the later requires significant computational resources to fine-tune. Thus, using Sheared Llama allows us to (i) run experiments where we fine-tune an LLM on in-domain data and (ii) share a model that is more useful to those in the community with limited access to computational resources. To compare two common inference scenarios with LLMs, we consider both settings: (i) \textbf{SL}: The out-of-the-box pre-trained Sheared LLaMA 1.3B model, and (ii) \textbf{SLFT}: A version of the Sheared LLaMA 1.3B model, fine-tuned on a curated dataset of 100,000 comments from r/AskReddit \footnote{\url{https://tinyurl.com/y6r68hkv}}. This model is fine-tuned using the Huggingface Transformers library\footnote{https://huggingface.co/} on an A100 GPU for 1 epoch, using a max token length of 128, a learning rate of $2e^{-5}$, and a batch size of 6. 

\section{Results and Implications}





Figure \ref{fig:preview} displays the result of our analysis. By plotting all scores of the human answers by the answer perplexity assigned by the \textit{SLFT} model, we can see a general trend. Across all 7,534 answers, the LLM models highly-rated answers better than poorly-rated answers. This indicates that the LLM is in line with human perception: it is probabilistically more likely to understand quality answers to fact-driven social media questions posed on r/Ask\{Topic\} communities. 

In Figure \ref{fig:main}, we see these same results of our analysis, separated by topic and model. Our fine-tuning strategy decreases LLM's perplexity across the board, indicating that fine-tuning helps the model understand human answers better. However, the general trend of the LLM modeling highly-rated answers better than poorly-rated answers exists for both the fine-tuned and vanilla LLMs. Similarly, we see the same trend across all individual subreddits from each of the three categories. We also find that LLMs models answers to fact-driven questions from \textit{professional} subreddit topics better than questions from \textit{social identity} or \textit{location} topics.


\paragraph{Implications for Future Work:}
In addition to the above-mentioned results, we highlight interesting insights surfacing from the data by considering different combinations of peer-assigned score of the answer and the level of LLM's perplexity for that answer.

\textbf{High Peer-assigned Score, High Perplexity}: LLMs fail to model these types of answers, and yet they are rated highly by humans.  As an example of this phenomenon, take the r/AskNYC question ``We have larger issues, but would anyone like to talk about silverfish?'' A highly-rated answer ``They are awful and it is alarming how fast they move'' has a high perplexity under the \textit{SLFT} model of 113.8, indicating the model hasn't learned the kind of language used in the answer even though the community found the answer appropriate for the question. Further exploring these types of answers might identify underrepresented answers compared with the ``typical'' answers found in LLM pre-training/fine-tuning data as well as potential blindspots of LLMs.

\textbf{Low Peer-assigned Score, Low Perplexity}: LLMs intuitively model these types of answers. However, even though they are in response to questions with high engagement, these types of answers are rated poorly by humans. Take this question from r/AskWomen ``Ladies, what's a cheap place to buy shoes?'' In response to this question, the answer ``Honestly I'd rather have a few pairs of quality shoes than lots of cheap ones'' may seem fairly reasonable; indeed, this answer is well-understood by the \textit{SLFT} model with a low perplexity of 7.8. However, this answer was poorly-rated, perhaps being seen as a divergent opinion from community consensus. Further exploration into these types of answers may lead to interesting analysis of divergent opinions on social media which may seem normal on a surface level.

In addition, future work could use this data and analysis framework to explore targeted fine-tuning of LLMs for socio-technical tools. Downstream work could use these tools to investigate similar questions in other domains and subreddits. In addition to these, we imagine several other hypotheses could be investigated with the data and analysis framework we present in this paper. We release our filtered dataset of these r/Ask\{Topic\} questions and answers\footnote{https://huggingface.co/datasets/pkseeg/reddit-ask-v0}, both fact-driven and subjective, as a rich source of hypotheses for future social science and web research.

    
    


\bibliography{refs}


\end{document}